\documentclass[10pt,twocolumn,letterpaper]{article}

\usepackage[pagenumbers]{cvpr} 

\usepackage{graphicx}
\usepackage{amsmath}
\usepackage{amssymb}
\usepackage{booktabs}
\usepackage{enumitem}
\usepackage{float}

\usepackage[pagebackref,breaklinks,colorlinks]{hyperref}

\usepackage[capitalize]{cleveref}
\crefname{section}{Sec.}{Secs.}
\Crefname{section}{Section}{Sections}
\Crefname{table}{Table}{Tables}
\crefname{table}{Tab.}{Tabs.}

\def\cvprPaperID{29} 
\def\confName{CVPR}
\def\confYear{2022}

\begin{document}

\title{medXGAN: Visual Explanations for Medical Classifiers through a Generative Latent Space}

\author{Amil Dravid$^{*1}$, Florian Schiffers$^{1}$, Boqing Gong$^{2}$, Aggelos K. Katsaggelos$^{1}$\\
\small{$^{1}$Northwestern University $^{2}$Google}\\
\tt\small  $^{*}$amildravid2023@u.northwestern.edu}

\maketitle

\begin{abstract}
   Despite the surge of deep learning in the past decade, some users are skeptical to deploy these models in practice due to their black-box nature.
   %
   %
   Specifically, in the medical space where there are severe potential repercussions, we need to develop methods to gain confidence in the models' decisions.
   To this end, we propose a novel medical imaging generative adversarial framework, medXGAN (\textbf{med}ical e\textbf{X}planation \textbf{GAN}), to visually explain what a medical classifier focuses on in its binary predictions.
   %
   %
   By encoding domain knowledge of medical images, we are able to disentangle anatomical structure and pathology, leading to fine-grained visualization through latent interpolation.
   Furthermore, we optimize the latent space such that interpolation explains how the features contribute to the classifier's output. 
   Our method outperforms baselines such as Gradient-Weighted Class Activation Mapping (Grad-CAM) and Integrated Gradients in localization and explanatory ability.
   Additionally, a combination of the medXGAN with Integrated Gradients can yield explanations more robust to noise. 
   The code is available at: \href{https://avdravid.github.io/medXGAN_page/}{https://avdravid.github.io/medXGAN\_page/}.
\end{abstract}

\section{Introduction}
\label{sec:intro}

Convolutional neural networks (CNNs) have enabled extremely accurate classification on large, complex datasets.
The ImageNet Large Scale Visual Recognition Challenge (ILSVRC)~\cite{russakovsky2015imagenet} kickstarted an era of massive efforts in tuning and finding new CNN architectures to beat classification benchmarks, among other tasks.
\begin{figure}[h!]
    \centering
    \includegraphics[width = 3.25in]{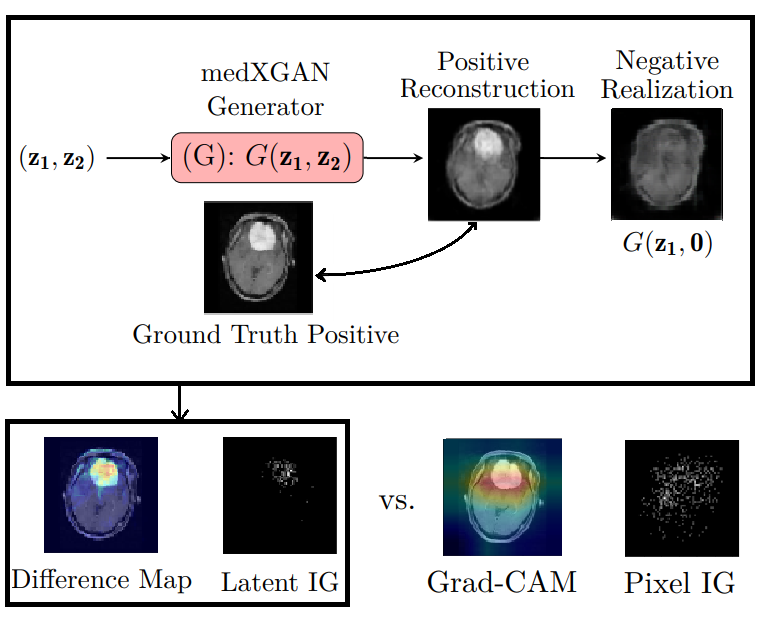}
    \caption{\textbf{Overview.} We propose a GAN framework medXGAN that takes in two latent vectors $(\mathbf{z1},\mathbf{z2})$ to encode anatomical structure and classifier-specific pathology, respectively. After training with a fixed classifier providing feedback, the generator can be used to explain the classifier's decision. Given a ground truth positive image, the latent code can be found via an optimization scheme. The positive image then can be turned into a negative realization by relying on the medXGAN's latent sampling scheme. The classifier-specific features can be visualized via a pixel-wise difference between the negative and positive images or by integrating the gradients (IG) by traversing the latent space (LIG). }
    \label{overview}
\end{figure}

Despite their performance, neural networks are largely considered to be black boxes by the machine learning community~\cite{alain2016understanding}.
As the size of these networks scale up with over millions of parameters~\cite{he2016deep}, this black box becomes even more complex. Although they demonstrate strong performance on artificially set-up tasks on datasets such as ImageNet\cite{russakovsky2015imagenet} among others, neural networks have been found to be extremely sensitive.
For instance, they perform poorly on data that is out-of-distribution with respect to their training set~\cite{zhang2021understanding, recht2019imagenet}.
Additionally, they break during inference on adversarial examples~\cite{goodfellow2014explaining}. Adversarial examples are images from the distribution that have visually imperceptible perturbations that drastically change the classifier's output.
These sensitivities drive the skepticism for deploying these models in actual practice. 

Particularly, there are tremendous consequences in the medical domain.
For example, with the onset of the COVID-19 pandemic, a slew of CNN models were created for COVID classification~\cite{alyasseri2021review}.
However, it has been found that many of them have been trained on biased datasets leading to a significant drop in performance on differently sourced datasets~\cite{degrave2021ai}.
These models were misled by visualization and validation techniques such as Gradient-Weighted Class Activation Mapping (Grad-CAM)~\cite{selvaraju2017grad}.
Following the surge of deep learning, the community has greatly increased efforts in explaining CNNs through various methods that we will explore later.
However, many lack the ability to localize with fine detail~\cite{8354201} or have even been found to be model-agnostic and fail to key in on the most important features~\cite{adebayo2018sanity}. 

Generative Adversarial Networks (GANs)~\cite{goodfellow2014generative}, a class of generative models, show promise in this task due to their ability to learn features and generate high fidelity images~\cite{ karras2019style}. Additionally, incorporating domain knowledge of the underlying data into the visualization shows promise in creating higher quality explanations~\cite{islam2021explainable}. As such, we propose a novel GAN framework, medXGAN (\textbf{med}ical e\textbf{X}planation \textbf{GAN}), to visually explain what a medical image-based CNN classifier has learned. This substantially builds upon our prior work~\cite{dravid2021visual}. Our scheme relies on encoding domain knowledge of medical images into the generator's latent sampling scheme while incorporating a pretrained classifier into the original GAN formulation. Given an image of the target class, we can find the latent representation and interpolate with the image's negative realization to visualize changing class features according to the CNN.

Our contributions are as follows:
\setlist{nolistsep}
\begin{itemize}[noitemsep]
    \item We propose the medXGAN framework that uses a classifier to explicitly disentangle the latent code into anatomical structure and classifier-specific features. There is no need to search in the latent space for these corresponding factors.
    \item We encode domain knowledge of medical images into the latent sampling scheme using a continuous class code to obtain desirable latent interpolation properties. 
    \item We propose using the negative realization of an image of the target class as a baseline for Integrated Gradients. We can then interpolate in the latent space, rather than pixel space, to obtain more localized and explanatory features. 
    \item We demonstrate the promise of our method over baselines such as Grad-CAM and Integrated Gradients in localization ability and explanatory power using both quantitative and qualitative experiments. 
\end{itemize}
\section{Background}


Generative Adversarial Networks (GANs) are a class of models that can generate new data from a target distribution ~\cite{goodfellow2014generative}. A GAN consists of a generator network (\emph{G}) and a discriminator network (\emph{D}) that are typically parameterized as neural networks. Their training scheme is analogous to an art forger trying to fool an art appraiser. The generator takes in a \emph{latent} or \emph{noise} vector $\mathbf{z}$ drawn from a random prior distribution $p_{z}$, such as a spherical normal distribution. From this, it tries to create images $G(z)$ that the discriminator classifies as real. However, the discriminator takes turns looking at real images ($x$) from the true distribution ($p_{data}$) and generated images $G(z)$ and tries to classify them as real or fake correctly.

The GAN objective is grounded in game theory through a minimax game with :
\begin{equation}
    \begin{aligned}
    \underset{G} {min} \text{ } \underset{D} {max} \text{ } &
    \underset{{\emph{x} \sim p_{x}}} {\mathbb{E}} [\log D(x)]\\
    & +\underset{{\textbf{\emph{z}} \sim p_{{\emph{z}}}({\emph{z}})}} {\mathbb{E}} [ \log (1- D(G({\emph{z}} )))]
    \end{aligned}
\end{equation}

The generator seeks to minimize the Jensen-Shannon (JS) divergence between its estimated distribution $p_{g}$ and the true distribution $p_{data}$.
The generator is an \emph{implicit density estimator}.
It learns to sample from a distribution rather than explicitly parameterizing it. The discriminator tries to minimize the divergence between its distribution $p_d$ and $p_{data}$ and maximizing the divergence of its estimated distribution for $p_g$ with $p_{data}$~\cite{goodfellow2016nips}. Equilibrium occurs when $p_g = p_{data}$ and the discriminator's output is $0.5$ for all images. 

The conditional GAN (C-GAN)~\cite{mirza2014conditional} is a natural extension, concatenating a discrete class code to the latent vector $z$ to control the generator's ability to synthesize images from different categories. The Auxiliary-Classifier GAN (AC-GAN) builds upon this by formulating the discriminator to output an auxiliary classification for input images~\cite{odena2017conditional}. Works such as~\cite{li2017triple, rangwani2021class} incorporate a separate classifier into the mix. Our work differs in that we include the classifier for the explicit objective of visualizing the classifier's learned attributes.

\subsection{The GAN's Latent Space}
The \emph{latent space} refers to a low dimensional space that captures factors of variation of the data, such as angle, pose, lighting, etc~\cite{kuhnel2018latent}. The generator learns to sample from this manifold and produce high fidelity images. This is done by sampling a latent vector \emph{\textbf{z}} drawn from some prior distribution. It has been found that interpolation and manipulation in this space can yield meaningful semantic results~\cite{DBLP:journals/corr/RadfordMC15, voynov2020unsupervised}.

Finding interpretable directions and learning representations that can separate informative factors of variations in the latent space is a highly active research topic~\cite{wang2020state}.

The task of finding a latent space consisting of linear subspaces controlling factors of variation is known as \emph{disentanglement}\cite{karras2019style}. Various GAN-based approaches have found success in unsupervised, supervised, and semi-supervised regimes~\cite{karras2019style, liu2020oogan, chen2016infogan,lin2020infogan,nie2020semi}. However, the corresponding factor for each subspace in these methods are arbitrary, and requires searching through them to find the factor of interest.    

\subsection{Visualization Methods}
The two most common traditional visualization methods in medical image include Gradient-Weighted Class Activation Mapping (Grad-CAM) and Integrated Gradients~\cite{selvaraju2017grad,sundararajan2017axiomatic, saporta2021benchmarking}. As such, we will focus on these two. 

\textbf{Gradient-Weighted Class Activation Mapping (Grad-CAM).} Gradient-weighted Class Activation Mapping
(Grad-CAM) relies on the gradients of a target concept  flowing into the final convolutional
layer, resulting in a coarse localization map. This highlights the regions that maximally activate the CNN for a particular class~\cite{selvaraju2017grad}. This map is known as a \emph{saliency map}. 

First, through backpropagation, the gradient of the score $y^c$ for class $c$ is calculated before the softmax with respect to feature maps $A^k$ of a convolutional 
layer $k$: $\frac{\partial y^c}{\partial A^k}$.

Next, these gradients are globally-averaged pooled to obtain $\alpha_{k}^c$, which are neuron importance weights describing the importance
of feature map $k$ for a target class $c$: Lastly, a ReLU function is applied to a weighted combination of feature maps and their corresponding neuron importance weights to obtain a positive-influence saliency map:

\begin{equation}
    L_{\text{Grad-CAM}}^c = ReLU\Bigg(\sum_k \alpha_{k}^{c}A^{k}\Bigg)
\end{equation}

A drawback of this method is its localization ability~\cite{8354201}.
Furthermore, it depends on the size of the convolutions. So, it tends to be biased towards larger models. In the medical domain, the lack of fine-grained detail can inadvertently capture a disease feature by the nature of "casting a wide net," leading to false confidence~\cite{wang2020score}. Additionally, this saliency map cannot tell the "whole story" and explain how the predicted features contribute to the prediction~\cite{schutte2021using}.

\textbf{Integrated Gradients (IG).}
Integrated Gradients (IG) relies on attributing the prediction of a deep network to its pixels of its input image~\cite{sundararajan2017axiomatic}.
Given a target image $x$ to visualize, a baseline image $x'$ is also established. There are many choices, but a completely black image is common~\cite{lundstrom2022rigorous}.
%
%
However, choosing an appropriate baseline image is an open problem~\cite{sturmfels2020visualizing}.
From there, a pixel-wise interpolation between these two images is fed into the classifier  $f$. The gradient is then taken with respect to the input pixels.
The parameter $\alpha$ governs the scale of interpolation.
As the interpolation from the black image approaches the target image, the gradients are accumulated and averaged. This leads to a map that highlights pixels that contain negative or positive attribution to the target class.
This is formulated as:
\begin{equation}
    \mathbf{IG} = (x-x') \int_{\alpha = 0}^{1} \frac{\partial f(x' + \alpha(x-x'))}{\partial x} d\alpha 
\end{equation}
although it discretized with summations in practice. 

Despite its ability to attribute importance at the pixel level rather than patch level as Grad-CAM does, Integrated Gradients is highly dependent on the chosen baseline~\cite{sturmfels2020visualizing}. Additionally, it can pick up noise and amount to an edge detector~\cite{adebayo2018sanity, drakard2021exploring}.

\textbf{Generative-Based Visualization.}
Generative models have been proposed to visualize classifiers~\cite{o2020generative, lang2021explaining, schutte2021using, mertes2020ganterfactual, li-2021-discover}. The work in~\cite{o2020generative} relies on Variational Autoencoders~\cite{kingma2013auto}, but is limited by experiments on artificial toy datasets. The methods in~\cite{lang2021explaining, schutte2021using} rely on StyleGAN~\cite{karras2019style} that generate high quality explanations, but lack substantial quantitative experiments on common baselines, and rely on search algorithms to find the relevant latent codes.

Our work is able to explicitly disentangle the latent code in a highly structured manner. Additionally, we show qualitative and quantitative experiments over the common baselines such as Grad-CAM and Integrated Gradients. The latent space in medXGAN is also optimized for meaningful latent interpolation that leads to the extension of Integrated Gradients in the latent space.

\section{Methods}
Utilitizing the medXGAN for feature visualization consists of three steps. First the classifier must be pretrained, and then incorporated into the training of the medXGAN. Then, given a ground truth positive image, a reconstruction task enables discovery of the latent vectors.
Lastly, the latent code can be used to generate a negative realization of the positive image. This yields powerful visualization capability as we can observe traverse the latent space to observe changing features, among other visualization methods.
\subsection{medXGAN Overview}
\begin{figure}[tp]
    \centering
    \includegraphics[width = 2.5in]{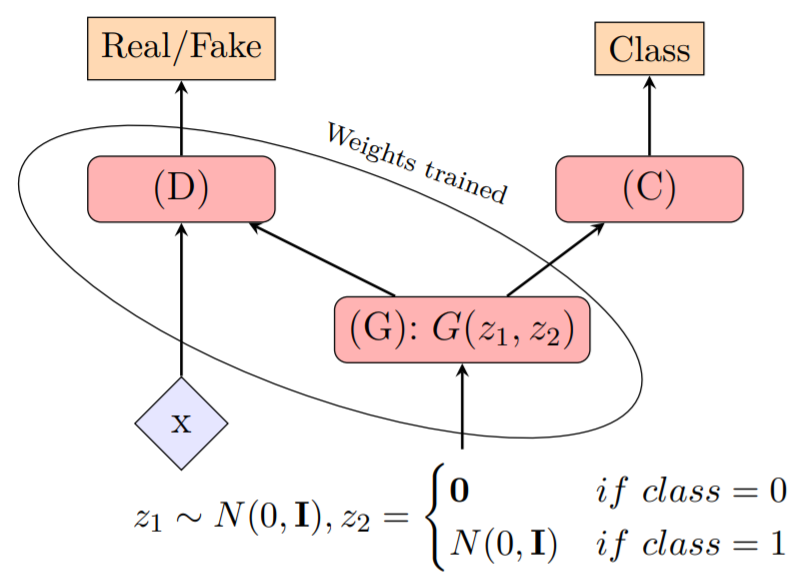}
    \caption{\textbf{medXGAN training scheme.} A pretrained classifier provides class feedback to the generator's synthesized image while the discriminator and generator play their typical adversarial game. The latent vector consists of $\mathbf{z_1}$ which encodes anatomical structure and $\mathbf{z_2}$ which corresponds to a continuous class code for pathology features according to the classifier.} 
    \label{medxgan}
\end{figure}

In order to visualize a CNN, we incorporate the pretrained network ($\emph{C}$) into the original GAN framework (see Fig.~\ref{medxgan}). The weights of the generator ($\emph{G}$) and discriminator ($\emph{D}$) are trained, playing the typical minimax game with real samples $\emph{x}$ and generated samples $G(\cdot)$. The weights for the classifier are fixed, thus this network provides feedback to the generator on the class (\emph{y}) according to the CNN's learned distribution $p_c$ . The overall objective is:
\begin{equation}
\begin{aligned}
    \underset{G} {min} \text{ } \underset{D} {max} \text{ } &
    \underset{{\emph{x} \sim p_{x}}} {\mathbb{E}} [\log D(x)]\\
    & +\underset {{z_1 \sim p_{z_1}, y \sim p_{y}}} {\mathbb{E}} [ \log (1- D(G(z_1, y )))]\\
    &-\underset {{z_1 \sim p_{z_1}, y \sim p_{y}}} {\mathbb{E}} [ \log ( p_c(y|G(z_1, y )))]
    \end{aligned}
    \label{medxganobjective}
\end{equation}
where the first two terms correspond to the original GAN formulation, and the third term relates to incorporating class features according to the CNN. The generator takes in two latent vectors that are concatenated. $z_1$ is drawn from a spherical normal distribution and corresponds to anatomical structure. $z_2$ corresponds to pathology features according to the classifier. If the image to be generated is negative (absent pathology) then $z_2$ is assumed to be $\mathbf{0}$. Otherwise, it is drawn from a spherical normal distribution.

\begin{figure}[t]
\captionsetup[subfigure]{labelformat=empty}
\centering

  
     \begin{subfigure}[t]{0.30\linewidth}
     \caption{{Label: 0 } }
     \caption{[1., 0.]}
     \includegraphics[height=\linewidth]{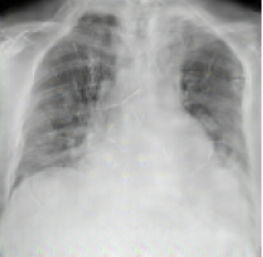}
     \end{subfigure}
     \hfill
     \begin{subfigure}[t]{0.30\linewidth}
     \caption{{Label: 1} }
     \caption{[0., 1.]}
     \includegraphics[height=\linewidth]{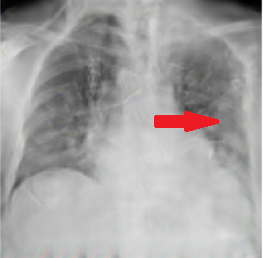}
     \end{subfigure}
     \hfill
     \begin{subfigure}[t]{0.30\linewidth}
     \caption{{Label: 1 } }
     \caption{[0., 1.]}
     \includegraphics[height=\linewidth]{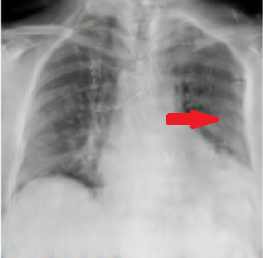}
     \end{subfigure}
  
    \caption{\textbf{Example of disentangled lungs and classifier features.} The classifier's softmax outputs are above along with label provided to generator. We can see that the lung and skeletal structure is intact, but features within the anatomy change, leading to different classifier outputs. The largest changes are highlighted by an arrow.
        }
    \label{disentangled}
\end{figure}
\begin{figure*}[h!]
\captionsetup[subfigure]{labelformat=empty}
\centering
  \begin{subfigure}[t]{0.12\textwidth}
  \caption{\hspace{0.1in}{Original}}
  \caption{{\hspace{0.1in}Positive Image}}
    \includegraphics[width=0.95in, height = 0.95in]{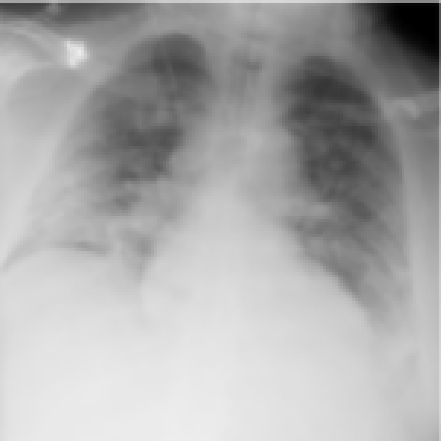}
     \caption{\hspace{0.13in}{[0., 1.]}}
  \end{subfigure}
     \hfill
  \begin{subfigure}[t]{0.12\textwidth}
  \caption{\hspace{0.1in}\normalsize{ } }
  \caption{\hspace{0.034in}{Reconstruction} }
    \includegraphics[width=0.95in, height = 0.95in]{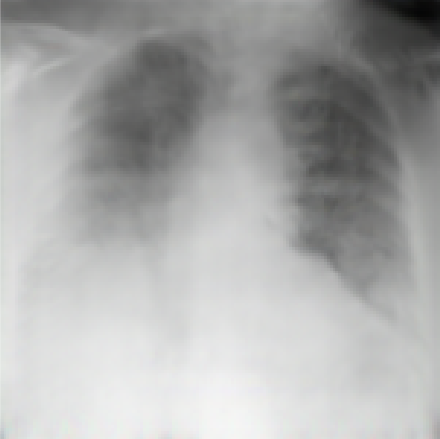}
    \caption{\hspace{0.13in}{[0., 1.]}}
  \end{subfigure}
     \hfill
  \begin{subfigure}[t]{0.12\textwidth}
  \caption{\hspace{0.1in}{Negative}}
  \caption{\hspace{0.1in}{Realization}}
    \includegraphics[width=0.95in, height = 0.95in]{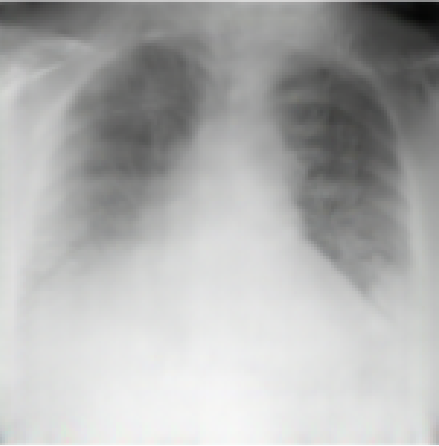}
    \caption{\hspace{0.13in}{[1., 0.]}}
  \end{subfigure}
     \hfill
  \begin{subfigure}[t]{0.12\textwidth}
  \caption{\hspace{0.1in}{Difference}}
  \caption{\hspace{0.1in}{Map}}
    \includegraphics[width=0.95in, height = 0.95in]{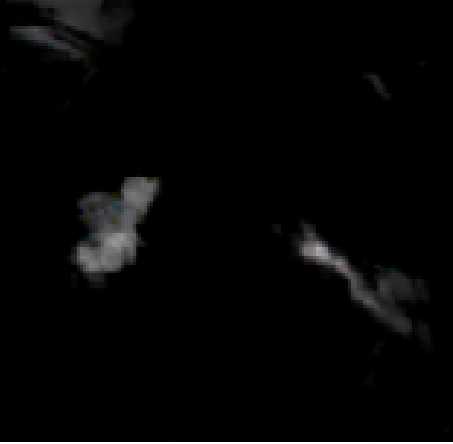}
  \end{subfigure}
     \hfill
  \begin{subfigure}[t]{0.12\textwidth}
  \caption{\hspace{0.1in}{Colorized}}
  \caption{\hspace{0.1in}{Map}}
    \includegraphics[width=0.95in, height = 0.95in]{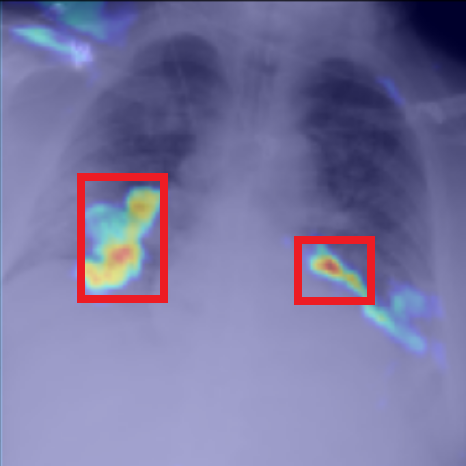}
    \end{subfigure}
    \hfill
  \begin{subfigure}[t]{0.12\textwidth}
  \caption{ }
  \caption{\hspace{0.1in}{Grad-CAM} }
  \vspace{0.01in}
    \includegraphics[width=0.95in, height = 0.96in]{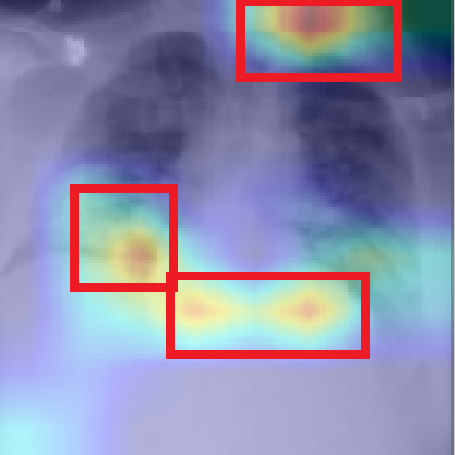}
  \end{subfigure}
  \caption{\textbf{Reconstructing a COVID positive image and turning it into negative}. Given a real positive class image, the latent code ($\mathbf{z_1},\mathbf{z_2}$) can be found via SGD with the generator to match pixel-wise and the classifier's output. The positive reconstruction can be turned into a negative realization by setting $\mathbf{z_2} = \mathbf{0}$. The classifier softmax outputs are below the respective images. Here, we visualize the pixel-wise difference between the realizations. Compared to Grad-CAM, we see more localization ability.}
  \end{figure*}
  \begin{figure*}[h!]
\captionsetup[subfigure]{labelformat=empty}
\centering
  \begin{subfigure}[t]{0.12\textwidth}
  \caption{\hspace{0.1in}{Baseline:}}
  \caption{\hspace{0.13in}{[1., 0.]}}
    \includegraphics[width=0.95in, height = 0.95in]{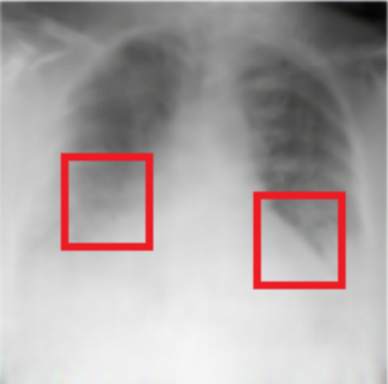}
  \end{subfigure}
     \hfill
  \begin{subfigure}[t]{0.12\textwidth}
  \caption{\hspace{0.1in}\normalsize{ } }
  \caption{\hspace{0.13in}{[0.88, 0.12]}}
    \includegraphics[width=0.95in, height = 0.95in]{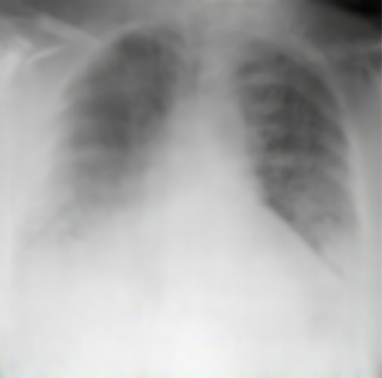}
  \end{subfigure}
     \hfill
  \begin{subfigure}[t]{0.12\textwidth}
  \caption{\hspace{0.1in}}
  \caption{\hspace{0.13in}{[0.56, 0.44]}}
    \includegraphics[width=0.95in, height = 0.95in]{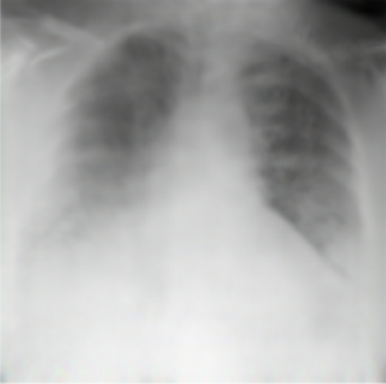}
  \end{subfigure}
     \hfill
  \begin{subfigure}[t]{0.12\textwidth}
  \caption{\hspace{0.1in}{}}
  \caption{\hspace{0.13in}{[0.47, 0.53]}}
    \includegraphics[width=0.95in, height = 0.95in]{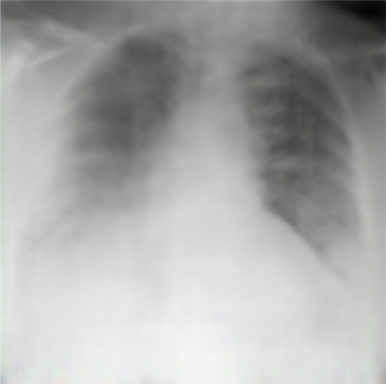}
  \end{subfigure}
     \hfill
  \begin{subfigure}[t]{0.12\textwidth}
  \caption{\hspace{0.1in}{}}
  \caption{\hspace{0.13in}{[0.15, 0.85]}}
    \includegraphics[width=0.95in, height = 0.95in]{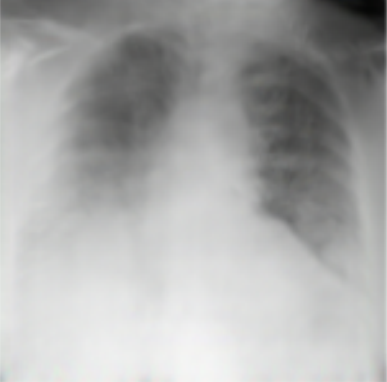}
    \end{subfigure}
    \hfill
  \begin{subfigure}[t]{0.12\textwidth}
  \caption{\hspace{0.1in}{Target}}
  \caption{\hspace{0.13in}{[0., 1.]}}
    \includegraphics[width=0.95in, height = 0.96in]{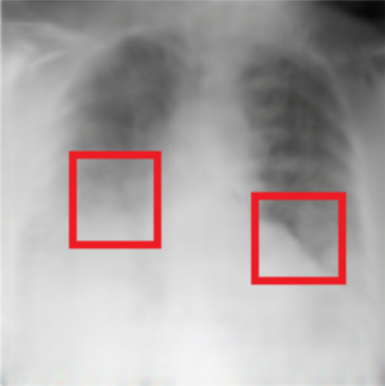}
  \end{subfigure}\\
  \vspace{0.05in}
  \begin{subfigure}[t]{0.12\textwidth}
    \includegraphics[width=0.95in, height = 0.95in]{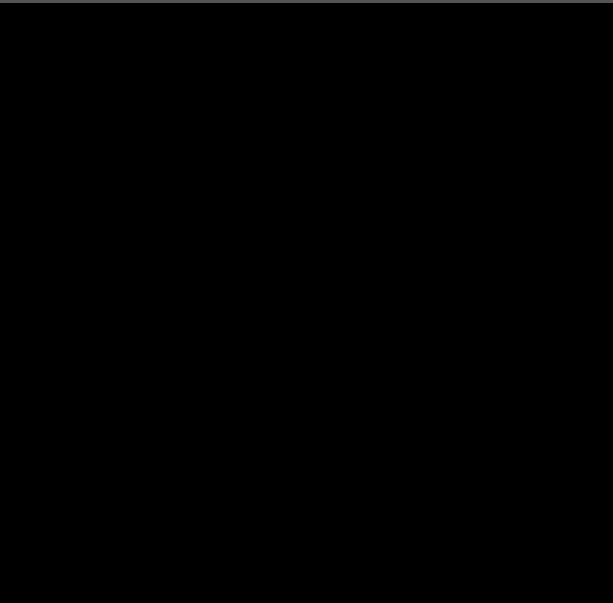}
  \end{subfigure}
     \hfill
  \begin{subfigure}[t]{0.12\textwidth}
    \includegraphics[width=0.95in, height = 0.95in]{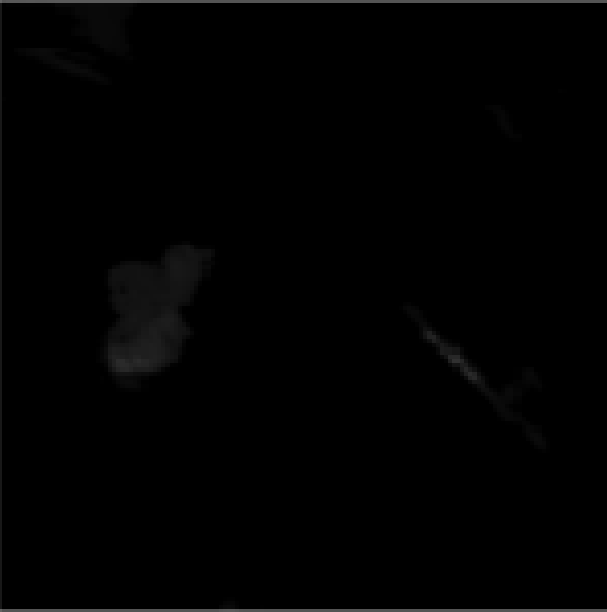}
  \end{subfigure}
     \hfill
  \begin{subfigure}[t]{0.12\textwidth}
    \includegraphics[width=0.95in, height = 0.95in]{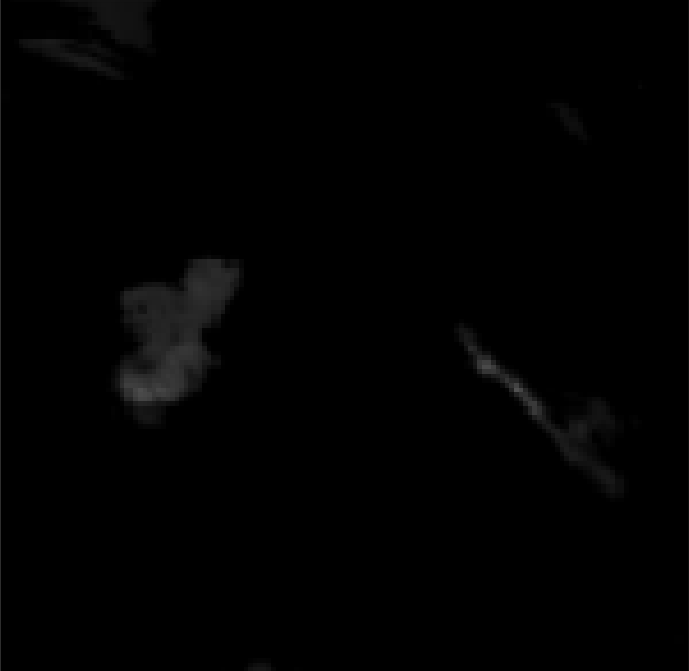}
  \end{subfigure}
     \hfill
  \begin{subfigure}[t]{0.12\textwidth}
    \includegraphics[width=0.95in, height = 0.95in]{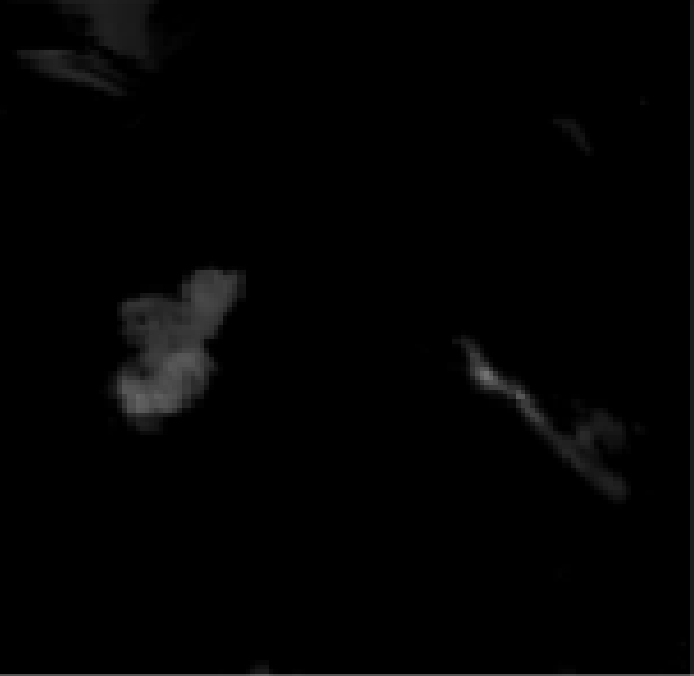}
  \end{subfigure}
     \hfill
  \begin{subfigure}[t]{0.12\textwidth}
    \includegraphics[width=0.95in, height = 0.95in]{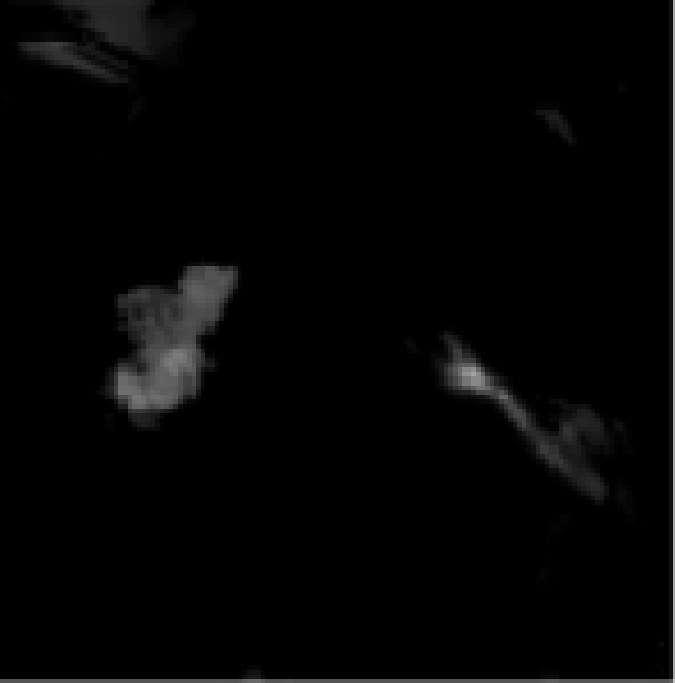}
    \end{subfigure}
    \hfill
  \begin{subfigure}[t]{0.12\textwidth}
    \includegraphics[width=0.95in, height = 0.96in]{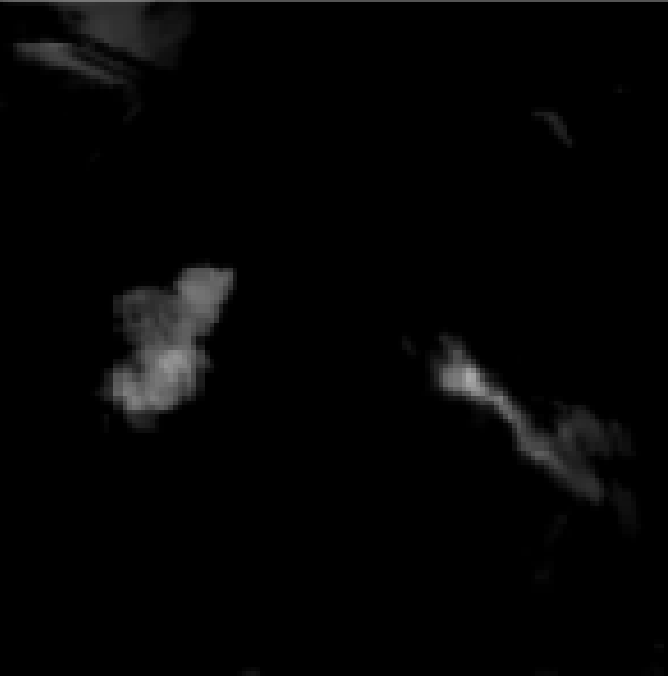}
  \end{subfigure}
  \caption{\textbf{Example of latent interpolation} Because $\mathbf{z_1}$ and $\mathbf{z_2}$ are disentangled latent codes, we can traverse the latent space by fixing $\mathbf{z_1}$ and stepping through $\mathbf{z_2}$. We observe fixed anatomical structure and changing pathology according to the classifier. The classifier's softmax outputs shown above each respective image. Additionally, we visualize the accumulating absolute value pixel-wise difference through the interpolation, which illustrates how the changing features contribute to the classifier's output.}
  \label{interpolation}
  \end{figure*}
After training the GAN, we can now visualize the CNN. Given a real positive image, we can find its latent representation via stochastic gradient descent for:
\begin{equation}
     \begin{aligned}
    \underset{z_1, z_2}{\text{arg min }}\text{MSE}(G(z_1,z_2),x) + \text{BCE}(C(G(z_1,z_2)), C(x))
    \end{aligned}
\end{equation}
where we are trying to match the pixels of the true image and reconstruction via a pixel-wise mean-squared error (MSE). We also match the classifier's output for both images with a binary cross-entropy loss(BCE). After we find $z_1$ and $z_2$, we can rely on the sampling scheme for $z_2$, changing it to $\mathbf{0}$ in order to convert the positive image to a negative reallization with high confidence while retaining the same anatomical structure.

Finally, we can interpolate in the latent space between the negative and positive images to visualize how the classifier's output changes with the interior pathology. We interpolate through the latent vector $z_2$ with steps $n$ at a rate of $\lambda$ while keeping the anatomical structure constant with $z_1$ by looking at the outputs of $G(z_1, \mathbf{0} + n\lambda z_2)$, for $n = 1, 2,...$
\subsection{A Disentanglement Perspective}
Mutual information describes the amount of information obtained about one random variable by observing the other random variable. Given two random variables, $(x,y)$, mutual information $I$ is related to entropy $H$: $I(x;y) = H(x)-H(x|y) = H(y) - H(y|x)$. It has been found that maximizing mutual information between some feature code $c$ and image $x$ can lead to disentangled representations\cite{chen2016infogan}. We can find a variational lower bound~\cite{agakov2004algorithm} for the mutual information $I(y;G(z_1,y))$. This method relies on the fact that the KL divergence between the posterior of the classifier's learned distribution and the true posterior is non-negative. 
\begin{equation}
\begin{split}
    I(y;G(z_1,y)) &= H(y) - H(y|G(z_1,y)) \\
    & \geq H(y) - \underset {{z_1 \sim p_{z_1}, y \sim p_{y}}}
    {\mathbb{E}} [\log( p_c(y|G(z_1, y )))]
\end{split}
\end{equation}
$H(y)$ can considered a constant term, so maximizing the mutual information between the class code and the generated image amounts to minimizing ${\mathbb{E}} [\log( p_c(y|G(z_1, y )))]$, which corresponds exactly to the third term of Eq.~\ref{medxganobjective}, thus leading to disentangled representations. We can see an example of this in Fig.~\ref{disentangled}.
\subsection{A Manifold Perspective}
According to the \emph{manifold hypothesis}, high dimensional data lies on lower dimensional manifolds in this space~\cite{fefferman2016testing}. However, natural data lies on a union of disjoint manifolds, and GANs struggle to model a distribution supported on disconnected manifolds\cite{khayatkhoei2018disconnected}. Interpolating between samples on disjoint manifolds may result in off-manifold samples. The Conditional-GAN induces disconnectedness by using a discrete code. In our case, we want "semantically smooth" interpolation in the latent space, with the classifier's output monotonically increasing as we interpolate from a negative and positive realization of a medical image. This lends itself to a smooth Integrated Gradients visualization that does not pick up on spurious features. We also want clinically plausible, on-manifold, intermediate results. As such we propose a continuous code that encodes domain knowledge of medical images. Typically there is an underlying anatomical structure that is fixed, but the disease pathology is not deterministic, and can manifest in multiple ways within the anatomy. As such, there is one realization of the negative image with $z_2 = \mathbf{0}$, and multiple realizations of a positive image with $z_2\sim\mathcal{N}(0,\mathbf{I}).$

\begin{figure} [!h]
    \centering
    \includegraphics[width = 3in]{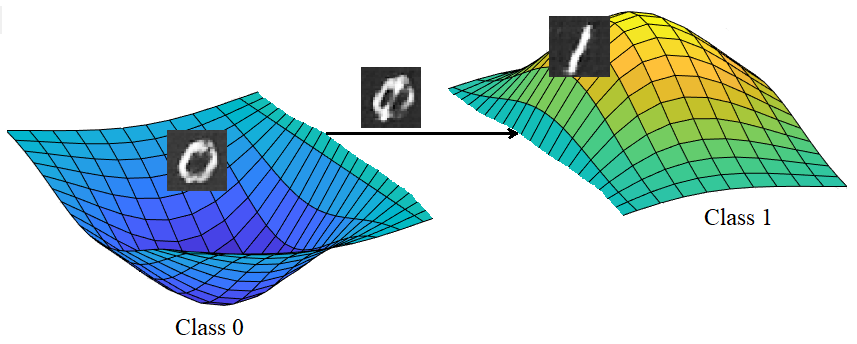}
    \caption{\textbf{Disconnected manifolds.} Interpolating between disconnected manifolds can lead to off-manifold intermediate results, which does not lend itself to a smooth and meaningful latent interpolation. For instance, the intermediate result for interpolating between 0 and 1 does not make natural sense. With medical images, we want a smooth transition with clinically plausible images.} 
\end{figure}

\section{Experiments}
We present qualitative analysis as well as quantitative experiments of our medXGAN method against the most popular explanatory techniques of Grad-CAM and Integrated Gradients. To begin, we first measure how well the generator captures the classifier's distribution. We first trained a VGG-16 network~\cite{simonyan2014very} to classify COVID-19 on an in-house dataset of COVID chest X-rays resized to $128\times128$~\cite{wehbe2021deepcovid}.
This network achieves roughly $75.0\%\pm0.9$ accuracy on the dataset, which was upsampled to become class balanced. Additionally, we trained an off-the-shelf CNN for binary classification of the presence of various brain tumor types on $64\times64$ MRIs~\cite{data}, achieving $85.20\%\pm1.3$ accuracy. The area under the receiver-operator (AUC) score for these is roughly of the same magnitude as the accuracy.  

We generated 4 images using the same anatomical structure by keeping $z_1$ fixed, with 1 negative and 3 positive realizations using the $z_2$ sampling scheme. This was repeated 1000 times to generate 4000 total images which were fed into the respective classifier for classification. For the MRI dataset, the classifier correctly predicts the class given to the generator with accuracy $\mathbf{99.2}\% \pm 0.2$. For the COVID dataset, the accuracy is $\mathbf{93.7}\% \pm 0.3$. Thus, there is a strong correspondence between the generator and classifier's distributions as the generator is incorporating classifier-specific features.  

\begin{table}
        \begin{tabular}{llll}\toprule
            Data && Accuracy & AUC \\\midrule
            MRI & Generated & $\mathbf{99.2}\% \pm 0.2$ & $\mathbf{0.995}\pm 0.001$ \\
                 & Real & $85.2\%\pm1.3$             & $0.975\pm 0.005$ \\
           X-Ray & Generated & $\mathbf{93.7}\% \pm 0.3$ & $\mathbf{0.980}\pm 0.003$ \\
                  & Real &  $75.0\%\pm0.9$             & $0.763\pm 0.008$
            \\\bottomrule
        \end{tabular}
        \caption{\textbf{Accuracy and AUC of classifier for tumor and COVID-19 classification on true validation images and generated data.} The high accuracy on the generated data suggests that the medXGAN generator is fitting strongly to the classifier's learned distribution based on its training data.}\label{Tab1}
    \end{table} 

\subsection{Grad-CAM Experiment}
Many experiments in the explainable AI space rely on counterfactual reasoning: "what would happen if we changed this feature?" Along these lines, we use Grad-CAM to create saliency maps for both brain MRIs and chest X-rays. Additionally, we use medXGAN to reconstruct negative and positive realization of the images, and take a pixel-wise difference to highlight the important changing features. After localizing the features through the two methods, we perturb the salient features and observe the change in the classifier's output on these new images, a commonly employed metric~\cite{poppi2021revisiting,o2020generative}. Although we can use perturbations such as Gaussian noise, or black or white pixels, we opt to replace the salient pixels with the average intensity of the image, and observe the average drop in the classifier's "positive" softmax output for multiple images. Given the grayscale images, black or white pixels may bias the decision of the model towards a particular class. Additionally, the output is sensitive to the particular instance of noise. For a fair evaluation with Grad-CAM, we do this for the medXGAN features as well instead of taking the negative realization of the positive image. The results of the counterfactual experiments are summarized in Tab.~\ref{counterfactualtable}, which indicates that the medXGAN is able to identify features with more explanatory power due to the greater drop in the classifier's output.
\begin{figure}
\centering
\captionsetup[subfloat]{labelformat=empty}
\subfloat[Original Positive]{\includegraphics[width=2.7cm, height = 2.7cm]{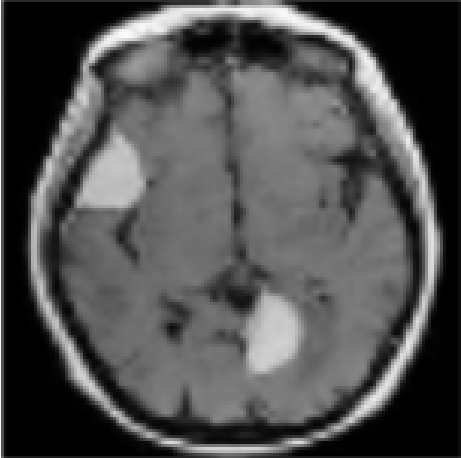}}\hfill
\subfloat[Reconstruction]{\includegraphics[width=2.7cm, height = 2.705cm]{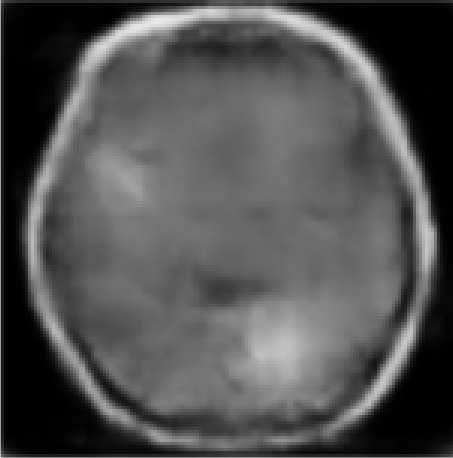}}\hfill
\subfloat[Negative Realization]{\includegraphics[width=2.7cm, height = 2.7cm]{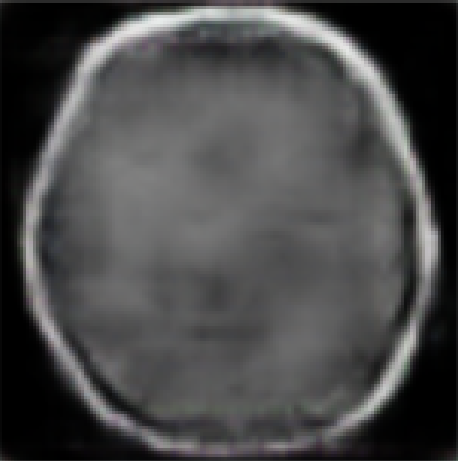}} \hfill
\subfloat[Difference Map]{\includegraphics[width=2.7cm, height = 2.7cm]{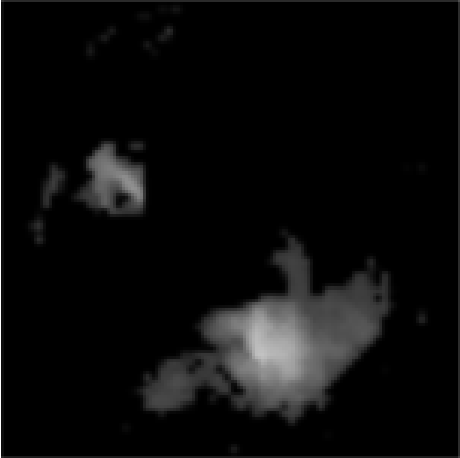}}\hfill  
\subfloat[Colorized Map]{\includegraphics[width=2.7cm, height = 2.7cm]{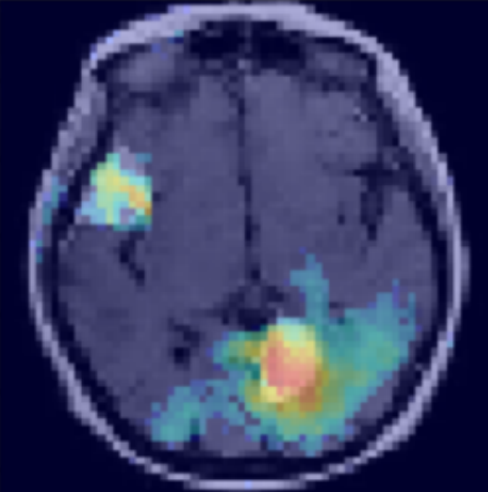}}\hfill
\subfloat[Grad-CAM]{\includegraphics[width=2.7cm, height = 2.70cm ]{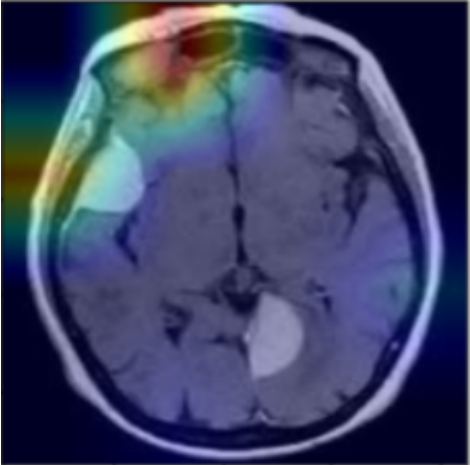}}
\caption{\textbf{Example of Grad-CAM and medXGAN methods on Brain MRIs for feature visualization.} In this example, despite lacking a perfect reconstruction, the medXGAN method localizes two tumors, while Grad-CAM focuses on one tumor and an eye.}
\label{figure}
\end{figure}

\begin{figure}[tb]
\centering
\begin{subfigure}[b]{\linewidth}
    \includegraphics[width=\linewidth, height = 0.9in]{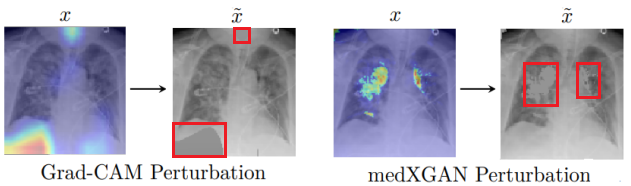}
    \caption{Perturb Features.}
  \end{subfigure}
  \begin{subfigure}[b]{\linewidth}
    \includegraphics[width=2.9in, height = 1.85in]{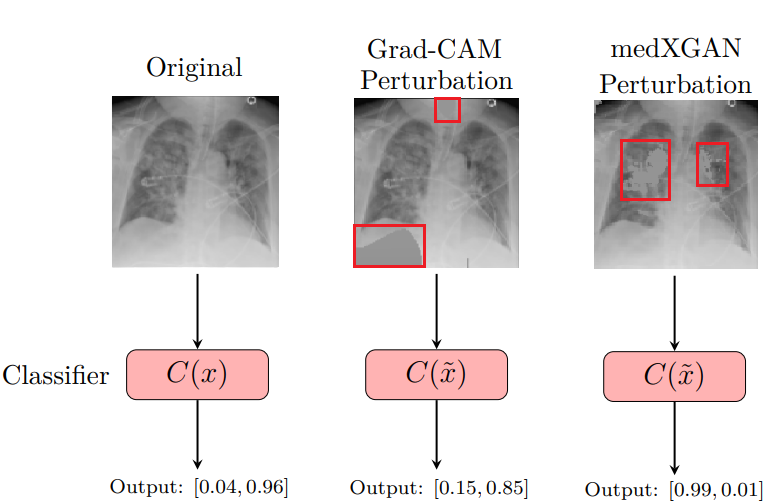}
    {\caption{Input into classifier.}}
  \end{subfigure}
  \caption{\textbf{Example of counterfactual perturbations}. We first feed a positive image from the validation set into a classifier and observe the classifier's output. Then, given the features highlighted from medXGAN and GradCAM, we perturb them to the mean intensity value of the original image. We then feed them into the classifier to observe the drop in confidence for the positive class. In this example, we see a larger drop in the classifier's output from the medXGAN method.}
  \end{figure}
  
\begin{table}[tb]
\centering
\begin{tabular}{l|ll}
         & MRI & X-Ray \\ \hline
medXGAN  &  $\mathbf{0.97}\pm0.03$     &  $\mathbf{0.91}\pm0.08$   \\
Grad-CAM &   $0.89\pm0.10$     &    $0.83\pm0.15$
\end{tabular}
\caption{\textbf{Results for counterfactual experiments}. We perturb features identified by medXGAN and Grad-CAM by changing them to the average intensity of the image. After feeding the image into the classifier, we observe the average drop in softmax output for positive. The medXGAN sees a larger drop in classifer confidence, suggesting that its identified features correspond more strongly to the classifier's decisions.}
\label{counterfactualtable}
\end{table}

\begin{figure*}[tb]
\captionsetup{justification=centering}

\centering
  \begin{subfigure}[t]{0.12\textwidth}
    \includegraphics[width= 0.82in,height = 0.82in]{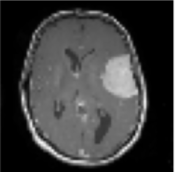}
  \end{subfigure}
     \hfill
  \begin{subfigure}[t]{0.12\textwidth}
    \includegraphics[width= 0.82in,height = 0.82in]{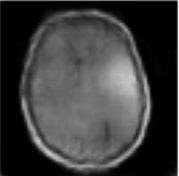}
  \end{subfigure}
     \hfill
  \begin{subfigure}[t]{0.12\textwidth}
    \includegraphics[width= 0.82in,height = 0.82in]{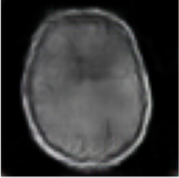}
  \end{subfigure}
     \hfill
  \begin{subfigure}[t]{0.12\textwidth}
    \includegraphics[width= 0.82in,height = 0.82in]{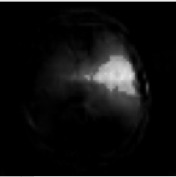}
  \end{subfigure}
     \hfill
  \begin{subfigure}[t]{0.12\textwidth}
    \includegraphics[width= 0.82in,height = 0.82in]{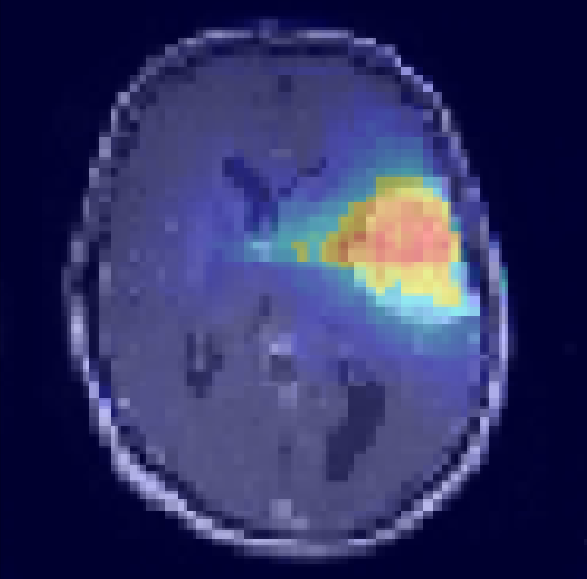}
    \end{subfigure}
    \hfill
  \begin{subfigure}[t]{0.12\textwidth}
    \includegraphics[width= 0.82in,height = 0.82in]{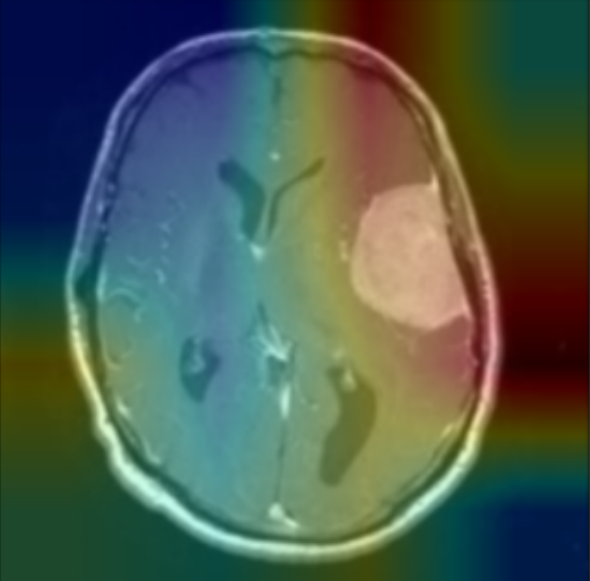}
  \end{subfigure}
     \hfill
  \begin{subfigure}[t]{0.12\textwidth}
    \includegraphics[width= 0.82in,height = 0.82in]{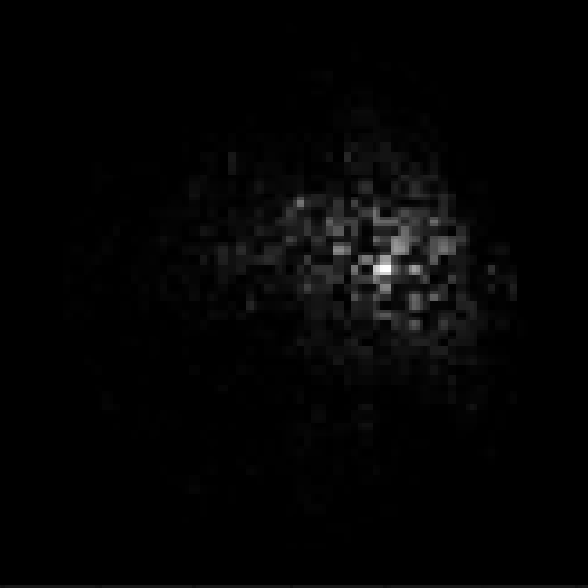}
  \end{subfigure}
      \hfill
  \begin{subfigure}[t]{0.12\textwidth}
    \includegraphics[width= 0.82in,height = 0.82in]{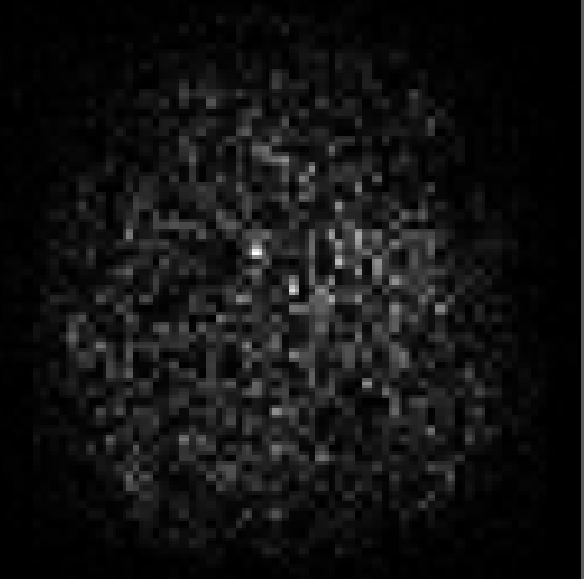}
  \end{subfigure}
  \begin{subfigure}[t]{0.12\textwidth}
    \includegraphics[width= 0.81in,height = 0.82in]{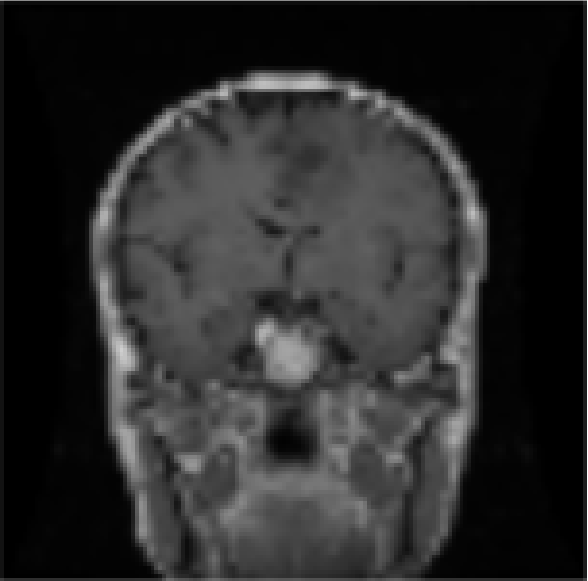}
    \caption{Original}
  \end{subfigure}
     \hfill
  \begin{subfigure}[t]{0.12\textwidth}
    \includegraphics[width= 0.82in,height = 0.82in]{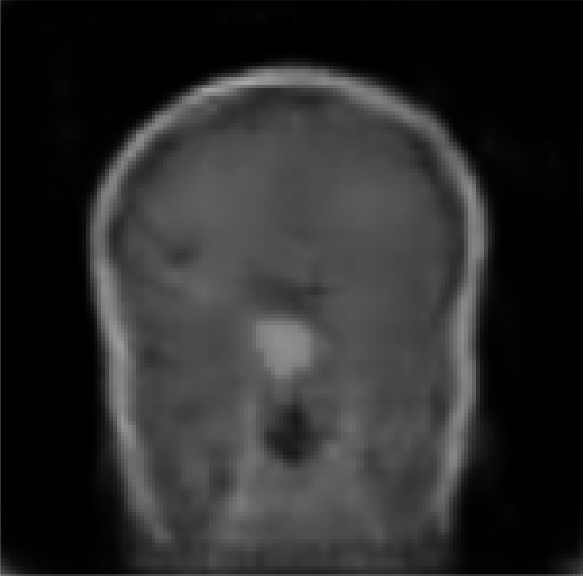}
    \caption{Reconstruction}
  \end{subfigure}
     \hfill
  \begin{subfigure}[t]{0.12\textwidth}
    \includegraphics[width= 0.82in,height = 0.82in]{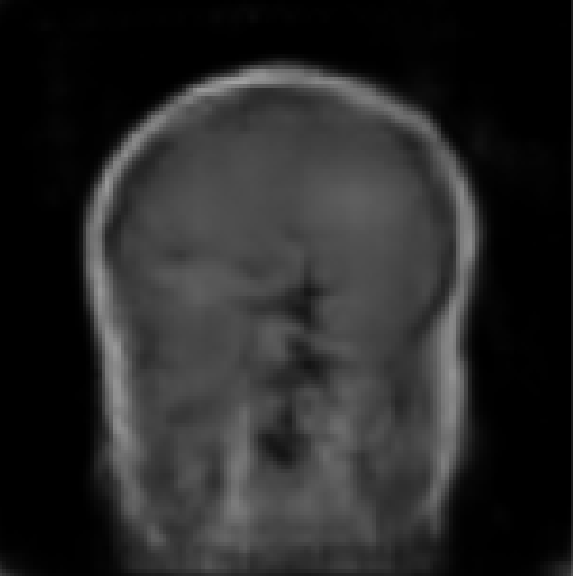}
    \caption{Negative Realization}
  \end{subfigure}
     \hfill
  \begin{subfigure}[t]{0.12\textwidth}
    \includegraphics[width= 0.82in,height = 0.82in]{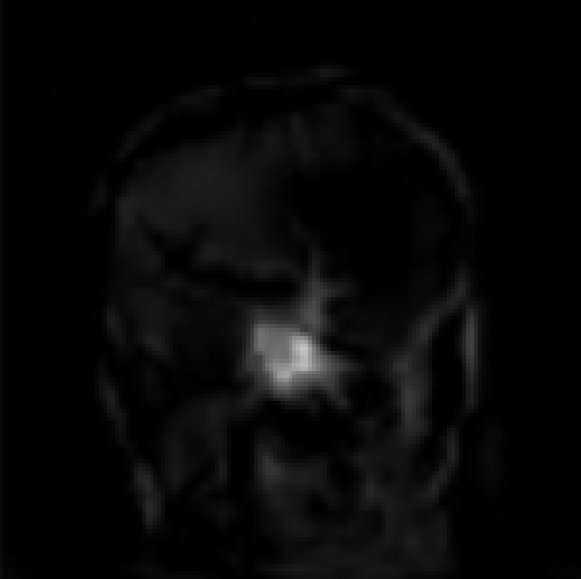}
    \caption{Difference Map}
  \end{subfigure}
     \hfill
  \begin{subfigure}[t]{0.12\textwidth}
    \includegraphics[width= 0.82in,height = 0.82in]{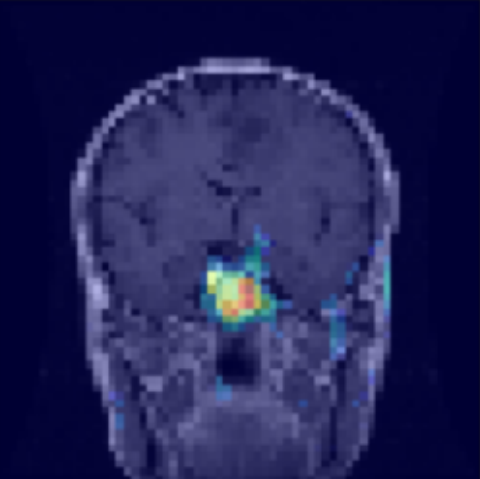}
    \caption{Colorized Map}
    \end{subfigure}
    \hfill
  \begin{subfigure}[t]{0.12\textwidth}
    \includegraphics[width= 0.82in,height = 0.82in]{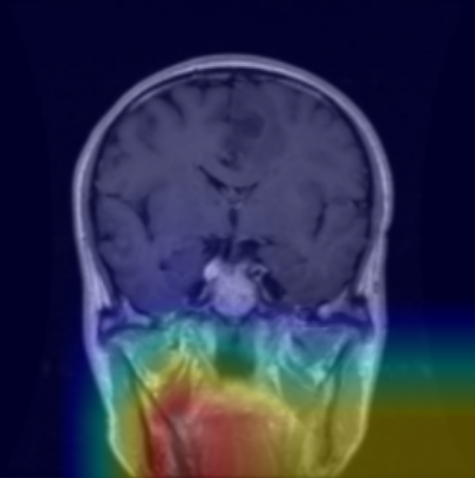}
    \caption{Grad-CAM}
  \end{subfigure}
     \hfill
  \begin{subfigure}[t]{0.12\textwidth}
    \includegraphics[width= 0.82in,height = 0.82in]{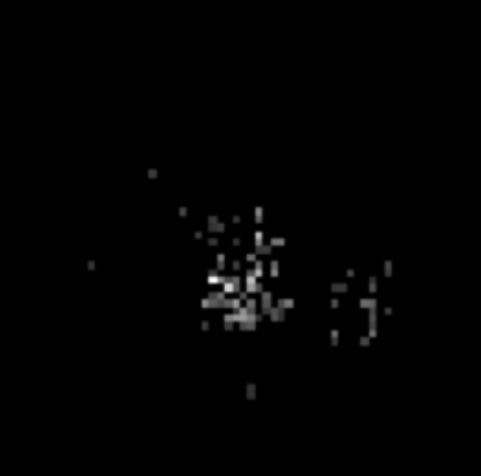}
    \caption{LIG}
  \end{subfigure}
     \hfill
  \begin{subfigure}[t]{0.12\textwidth}
    \includegraphics[width= 0.82in,height = 0.82in]{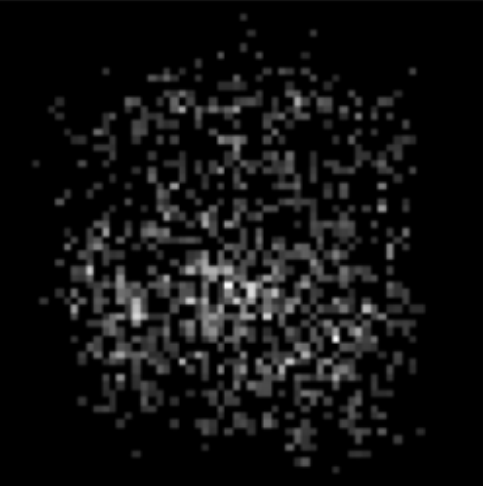}
    \caption{IG}
  \end{subfigure}
  
 \captionsetup{justification=justified}

  \caption{\textbf{Example of visualizations on the brain MRI dataset.} Both difference maps and integrating gradients through the latent space (LIG) localize important classifier-specific features much finer than Grad-CAM and standard pixel-wise Integrated Gradients (IG).}
  \label{tumorviz}
\end{figure*}
\subsection{Integrated Gradients Experiment}
As visualizing brain tumors is more interpretable to non-experts, we opt to use Integrated Gradients with just the brain MRI dataset. For this experiment we measure the  degree of localization. We apply the standard Integrated Gradients to MRIs with tumors by interpolating in the pixel space between a black image and the target image as is standard practice. Additionally, we propose to use our medXGAN to interpolate in the latent space between the negative and positive realizations of the image, which is one of our novel contributions. We refer to this as Latent Integrated Gradients (LIG). This can be formulated as:
\begin{equation}
\begin{aligned}
    & \mathbf{LIG} =\\
    &[G(z_1, z_2) - G(z_1, \mathbf{0})] \int_{\alpha = 0}^{1} \frac{\partial f(G(z_1, \mathbf{0} + \alpha z_2))}{\partial x} d\alpha 
\end{aligned}
\end{equation}
where were are taking the gradient with respect to the input $x$ to the model $f(x)$ as we interpolate in the latent space. Afterwards, we find the ratio of pixels with some non-zero attribution value through latent vs. pixel interpolation. The averaged ratio over multiple images was $\mathbf{0.21} \pm .09$, indicating that our method is able to localize the salient features with much finer detail. Essentially, it is using one-fifth the number of pixels that standard Integrated Gradients attributes. Our method does not capture as much noise or as many spurious edges as the baseline does~(Fig.~\ref{tumorviz}). 

\subsection{Qualitative Analysis}
In both the brain tumor and COVID dataset, we see that the medXGAN is able to attribute classifier-specific features that are fine-grained compared to Grad-CAM or Integrated Gradients. For instance, medXGAN is able to completely capure the brain tumor in Fig.~\ref{tumorviz} while Grad-CAM misses it. Although the reconstructed images are not of the highest fidelity to the ground truth, they still capture the important anatomical structure and pathological features, lending to successful visualization. Additionally, we notice that through latent interpolation, the classifier's output for positive is monotonically increasing, a result based on connecting two disconnected class manifolds through a continuous class code. While traditional visualization methods will give a single map highlighting salient regions, our method can be employed to study how the features contribute to the classifier's output through latent interpolation. Additionally, we implicitly get access to the classifier's decision boundary as we can observe when the classifier changes predictions based on the latent interpolation (Fig.~\ref{interpolation}).

\section{Convergence Experiments}
As visualization with the medXGAN relies on finding the latent code of ground truth images, we quantitatively examine how visualizations can change depending on different runs of the optimization scheme. We randomly initialize the latent vectors with values from a standard normal distribution. For the medxGAN trained on the brain MRIs, the latent vectors are $z_1\in\mathbb{R}^{1000\times1}$ and $z_2\in\mathbb{R}^{100\times1}$. For the GAN trained on chest X-rays, they are $z_1\in\mathbb{R}^{100\times1}$ and $z_2\in\mathbb{R}^{10\times1}$.  On various positive class images from the MRI and X-ray datasets, we run stochastic gradient descent on $(\mathbf{z_1}, \mathbf{z_2})$ multiple times for a fixed $10,000$ epochs. We then compute two metrics. First, we find the pairwise cosine similarity between the latent vectors found between the multiple runs. This measures the similarity in the latent space. We then run the vectors through the generator and measure the perceptual similarity via the Structural Similarity Index Measure (SSIM)~\cite{1284395}. These results are summarized in Table~\ref{convergence}. As these scores all tend towards 1, we see that despite the nonconvex optimization scheme, we are converging to very similar regions in the image and latent space. As the cosine similarity and SSIM are both approaching one, it appears that convergence in the latent space correlates to convergence in the image space. However, this can be further studied. 
\begin{table}[!h]
\centering
\begin{tabular}{l|ll}
         & Cosine Similarity & SSIM \\ \hline
MRI  &  $0.965\pm0.008$     &  $0.970\pm0.020$   \\
 X-Ray &   $0.994\pm0.005$     &    $0.968\pm0.015$
\end{tabular}
\caption{\textbf{Convergence results.} Based on random initial seeds, we optimize over the latent vectors for 10,000 epochs to reconstruct various images. We compute the SSIM between the respective images as well as the cosine similarity between the latent vectors. With the metrics approaching 1, the reconstructions seem to converge in the image and latent space. }
\label{convergence}
\end{table}
\begin{figure}
\centering
\begin{subfigure}[t]{0.30\linewidth}
    \includegraphics[width=\textwidth, height = 0.95in]{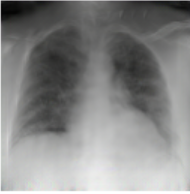}
  \end{subfigure}
  \hfill
  \begin{subfigure}[t]{0.30\linewidth}
    \includegraphics[width=\textwidth, height = 0.95in]{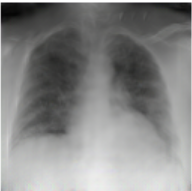}
  \end{subfigure}
    \hfill
  \begin{subfigure}[t]{0.30\linewidth}
    \includegraphics[width=\textwidth, height = 0.95in]{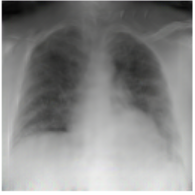}
  \end{subfigure}
  
\vspace{1mm}

\begin{subfigure}[b]{0.30\linewidth}
    \includegraphics[width=\textwidth, height = 0.95in]{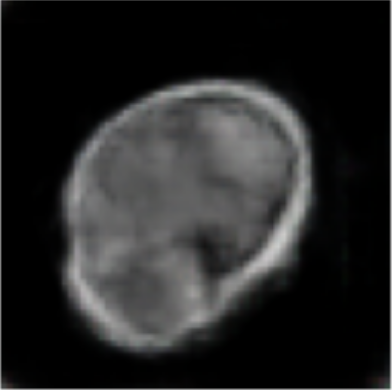}
  \end{subfigure}
    \hfill
  \begin{subfigure}[b]{0.30\linewidth}
    \includegraphics[width=\textwidth, height = 0.95in]{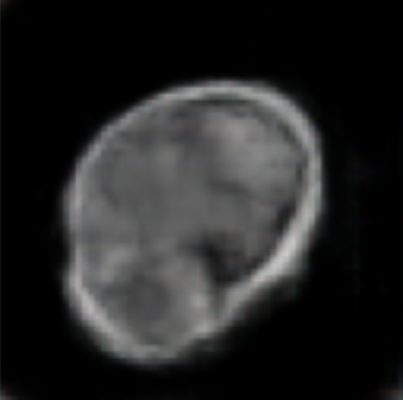}
  \end{subfigure}
   \hfill
  \begin{subfigure}[b]{0.30\linewidth}
    \includegraphics[width=\textwidth, height = 0.95in]{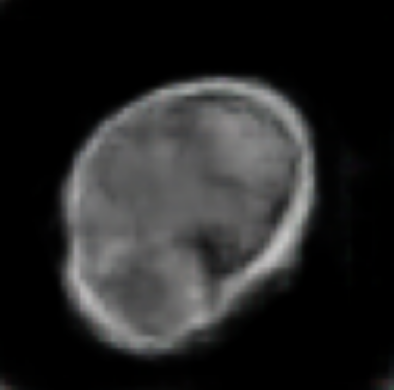}
  \end{subfigure}
  \caption{\textbf{Example of converging reconstructions.} Given random initial seeds for the latent vectors, the reconstructions appear to converge to extremely similar images.
  }
  \label{convergefig}
  \end{figure}
\section{Limitations and Discussion}
\begin{figure}
\centering
\captionsetup[subfloat]{labelformat=empty}
\begin{subfigure}[b]{0.12\textwidth}
 \includegraphics[width=\textwidth, height = 0.79in]{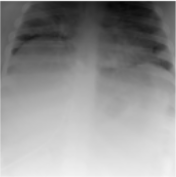}
 \caption{Original X-Ray}
  \end{subfigure}
  \begin{subfigure}[b]{0.11\textwidth}
    \includegraphics[width=\textwidth, height = 0.79in]{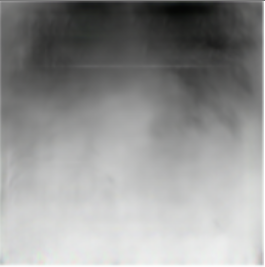}
    \caption{Reconstruction}
  \end{subfigure}
  \begin{subfigure}[b]{0.11\textwidth}
    \includegraphics[width=\textwidth, height = 0.81in]{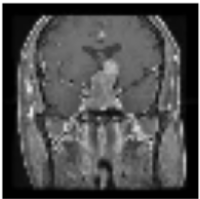}
    \caption{Original MRI}
  \end{subfigure}
\begin{subfigure}[b]{0.11\textwidth}
    \includegraphics[width=\textwidth, height = 0.80in]{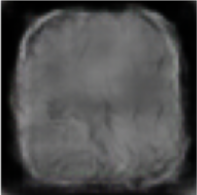}
    \caption{Reconstruction}
  \end{subfigure}
\caption{\textbf{Failed reconstructions.} These original images can be considered anomolies as some of the characteristics they present are not well represented in the dataset. For instance both the X-ray and MRI are heavily zoomed in. Although the reconstruction captures some structure, it misses many details within.}
\label{failed}
\end{figure}

Despite the promising results of medXGAN, we recognize some limitations of our method. One is based on training data. It is well-known that GANs require a significant amount of training data in order to faithfully "learn" the training distribution~\cite{noguchi2019image}. In our experiments, the generation of chest x-rays are much higher fidelity than the brain MRIs due to the dataset sizes: $\sim 15000 \text{ vs.} \sim 2000$ images. Nevertheless, even with the brain MRIs, we are able to capture important classifier-specific features. To extend the generator's capacity given limited data, we suggest employing a data augmentation scheme such as~\cite{karras2020training} or a transfer learning approach like~\cite{zhao2020leveraging}. Additionally, faithful image reconstruction relies on the GAN learning a rich latent space. If the generator becomes too adapted to the training distribution, examples out-of-distribution or "off-manifold" may result in poor reconstructions~\cite{kang2021gan,webster2019detecting}. In Fig.~\ref{failed}, we see examples of poor reconstructions as the ground truth images are not well-represented in the distribution. We plan to scale up the GANs to higher resolution and more optimized frameworks for the highest fidelity image synthesis. However, we have seen that even rough approximations of the ground truth images can be very powerful. Additionally, given the restriction to binary classification, our next steps include extending to multi-class classification. 

Despite the growing field of explainable AI, quantitative benchmarking and comparison is still an open problem~\cite{das2020opportunities}. Evaluation can largely be \emph{ad hoc}. For instance, we determined that using the mean intensity of the image, would serve as the fairest perturbation without inducing bias~\cite{jain2021missingness}. We could also use an image inpainting technique instead~\cite{guillemot2013image}. We can further validate our method by measuring its reliance on the classifier's weights as well as the dataset labels through methods proposed in~\cite{adebayo2018sanity}.

Despite the computational overhead of training a GAN for visualization, the generator can also provide meaningful data augmentation to further optimize the classifier~\cite{shorten2019survey}. With the disentangled latent codes, users have more control over the generated images. Additionally, the latent space is optimized so that interpolation leads to a monotonic increase in the classifier's output. As such, this can be leveraged to create samples near the classifier's decision boundary. 

\section{Conclusion}
In this work, we presented medXGAN, a novel GAN framework that encodes domain knowledge of medical images into its latent sampling scheme through a continuous class code. This allows for explicit disentanglement of anatomical structure and classifier-specific pathology features. Additionally, we proposed using a negative realization of a positive class image as a baseline along with latent interpolation for Integrated Gradients. We establish this as Latent Integrated Gradients (LIG). We also demonstrated medXGAN's promising explanatory and localization power through quantitative and qualitative analysis over the baselines of Grad-CAM, Integrated Gradients. It is important to note that the visualizations are not what the actual class features should be, but rather what the classifier thinks. So the visualizations are subject to the biases and errors of the classifier. Ultimately, we hope our method inspires further efforts to open the black box of neural networks.
{\small
\bibliographystyle{ieee_fullname}
\bibliography{egbib}
}

\end{document}